\documentclass[lettersize,journal]{IEEEtran}
\usepackage{amsmath,amsfonts}
\usepackage{algorithmic}
\usepackage{algorithm}
\usepackage{array}
\usepackage[caption=false,font=normalsize,labelfont=sf,textfont=sf]{subfig}
\usepackage{textcomp}
\usepackage{stfloats}
\usepackage{url}
\usepackage{verbatim}
\usepackage{graphicx}
\usepackage{cite}
\usepackage[export]{adjustbox}
\usepackage{makecell}
\usepackage{pifont}
\newcommand{\xmark}{\ding{55}}%
\newcommand{\cmark}{\ding{51}}%
\usepackage{hyperref}
\usepackage{xcolor}
\usepackage{caption}
\usepackage{tablefootnote}
\usepackage{multirow}

\begin{document}

\title{BehavePassDB: Public Database for Mobile Behavioral Biometrics and Benchmark Evaluation}

\author{\IEEEauthorblockN{Giuseppe Stragapede,
Ruben Vera-Rodriguez, Ruben Tolosana, Aythami Morales}  \\
\IEEEauthorblockA{\textit{Biometrics and Data Pattern Analytics},\\
Universidad Autonoma de Madrid\\
Madrid, Spain\\
Email: giuseppe.stragapede@uam.es,
ruben.vera@uam.es,
ruben.tolosana@uam.es,
aythami.morales@uam.es}
}

\maketitle

\begin{abstract}
Mobile behavioral biometrics have become a popular topic of research, reaching promising results in terms of authentication, exploiting a multimodal combination of touchscreen and background sensor data. However, there is no way of knowing whether state-of-the-art classifiers in the literature can distinguish between the notion of user and device. In this article, we present a new database, BehavePassDB, structured into separate acquisition sessions and tasks to mimic the most common aspects of mobile Human-Computer Interaction (HCI). BehavePassDB is acquired through a dedicated mobile app installed on the subjects’ devices, also including the case of different users on the same device for evaluation. We propose a standard experimental protocol and benchmark for the research community to perform a fair comparison of novel approaches with the state of the art\footnote{\texttt{\url{https://github.com/BiDAlab/MobileB2C_BehavePassDB/}}}. We propose and evaluate a system based on Long-Short Term Memory (LSTM) architecture with triplet loss and modality fusion at score level.
\end{abstract}

\begin{IEEEkeywords}
mobile authentication, continuous authentication, behavioral biometrics, BehavePassDB, device bias
\end{IEEEkeywords}

\section{Introduction}
\label{sec:Introduction}
Mobile biometric authentication currently relies mostly on physiological biometrics such as fingerprint or face \cite{WANG2020107118}. These biometric systems, however, are prone to physical presentation attacks (spoofing) \cite{marcel2019handbook} and digital manipulations \cite{rathgeb2022handbook} and, just as well as knowledge-based systems (PIN codes, passwords, and lock patterns) \cite{8998358}), they are designed for \textit{entry-point} authentication and not suited for offering prolonged protection. In such case, mobile users would have to keep interrupting their activity to carry out the authentication process, for instance by placing their fingertip on the dedicated scanner. Frequent face verification also seems infeasible due to hardware constraints, such as the computational overload, memory overhead and battery consumption of the acquisition and processing of images. Considering such limitations of the currently deployed authentication systems, if an intruder gains access to the device, they can stay authenticated as long as the device remains active, being granted a considerable amount of time to obtain private information.
%\cite{Perera2017}.
\par In this scenario, in contrast to physiological biometrics, behavioral biometrics\footnote{In the context of biometrics, a well-known dichotomy is given by the nature of the traits: all biological characteristics that allow to identify an individual are defined as \textit{physiological} (face, fingerprint, iris, etc.), whereas all the means that allow or help in discriminating among individuals based on the \textit{way} activities are performed, such as gait, typing, scrolling, signature, etc., are labelled as \textit{behavioral}.} allow for Continuous Authentication (CA), a paradigm based on constantly verifying the biometric features of the user in a \textit{passive} way, in other words, without having them to carry out any specific authentication task \cite{7503170, stragapede2022prl}. In CA systems, biometric samples are continuously acquired and processed, and the user will be redirected to an \textit{entry-point} authentication mechanism in case that the matching with pre-acquired enrolment samples returns a negative response. To this end, behavioral biometrics traits are suitable as mobile devices are equipped with several sensors, such as touchscreen, motion sensors, etc., able to continuously acquire low-dimensional temporal signals concerning the user activity, that can reveal a significant amount of information about the user \cite{delgadosantos2021survey}. 
In addition, other aspects such as the application usage, GPS position, and network connections are included in the behavioral category as they capture users' personal habits and routines. As a result, behavioral biometrics offer strong security and high usability in this mobile scenario.

\subsection{Description of the Problem}
\label{subsec:Description_of_the_Problem}
\par Although systems based on behavioral biometrics do not usually achieve the same authentication performance as their \textit{physiological} counterparts, Behavioral Biometrics for Continuous Authentication (BBCA) \cite{7503170} is an appealing area for the biometrics research community, either as: 
\begin{itemize}
    \item[\textit{(i)}] A form of complementary technology or second factor in a 2-factor authentication (2FA) \cite{8998358}; for instance, in a remote security-wise critical service, BBCA could be used on top of existing security protocols. \textbf{In this case, every user would be using their own mobile device.} 
    \item[\textit{(ii)}] The primary security technology in the real-world scenario of a theft, in which the \textbf{impostor and the genuine user data originate from the same device} \cite{rasmussen15NDSS}. 
\end{itemize}

\begin{figure*}[t]
	\centering
% 	\hspace*{\linewidth}
    \includegraphics[width=\linewidth]{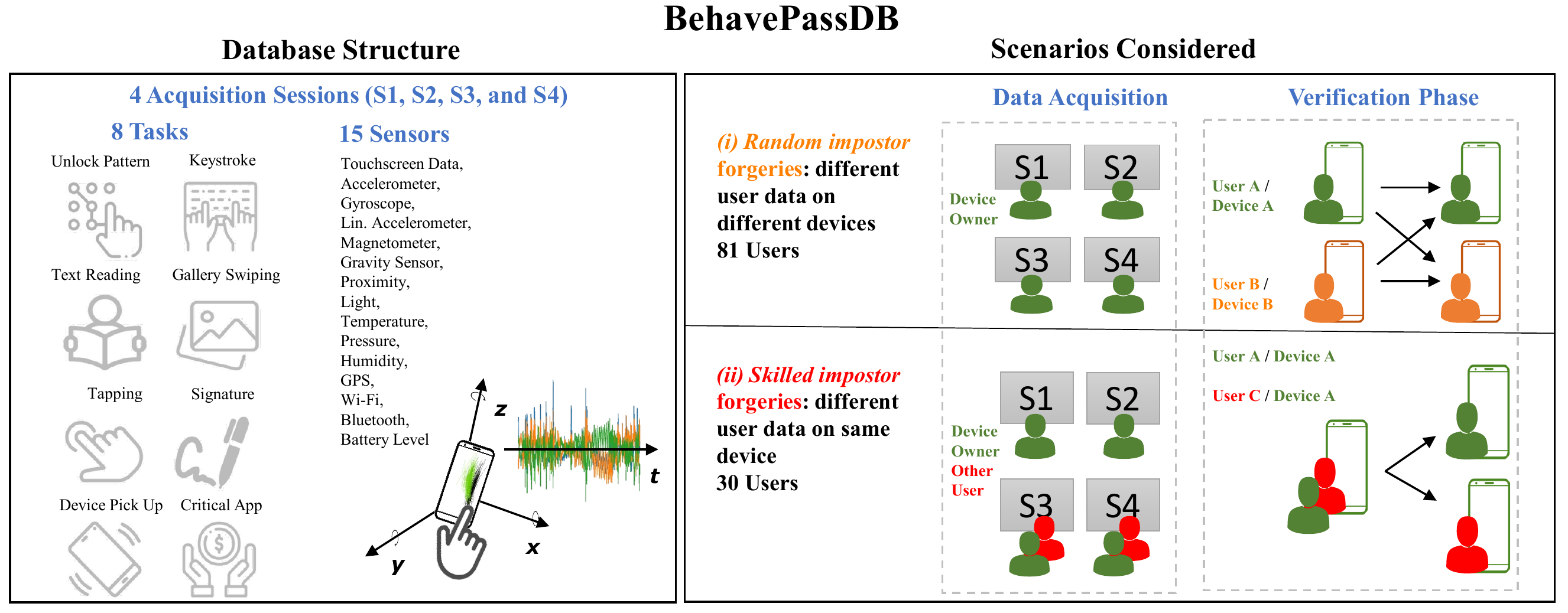}
	\captionsetup{width=\textwidth}
    \caption{In BehavePassDB, each of the 4 acquisition sessions contains 8 tasks. During each task, data are acquired from 15 different sources (modalities), including touchscreen information and background sensors. Two impostor scenarios are allowed for user verification: {\color{orange}\textit{(i)} \textit{random} impostor forgeries}, in which every user data are acquired from their own mobile device, and the user data cross comparisons entail different devices; {\color{red}\textit{(ii)} \textit{skilled} impostor forgeries}, in which the last two acquisition sessions are performed also by a different user on the same device imitating the legitimate one, allowing user data cross comparisons from the same device.}
    \label{fig:captures}
\end{figure*}

The difference in between the two scenarios consists in the authentication technology being able to differentiate between the notion of ``device” and ``user”.
This is mainly related to two data collection approaches, as subjects are typically asked to use their own mobile device for data collection without hard requirements in terms of the devices used, as long as it is the same one for the entire data acquisition process \cite{Acien2020b}, or they are asked to use only one or few dedicated acquisition devices \cite{Zhu2019}.
% Vajdi2019, Garbuz2019, Cilia2018}. 
Consequently, in the first case, the user extracted features of user might in reality be related to their mobile device, and it is not possible to assert with certainty that they are, indeed, biometric. A potential learning bias could be in fact introduced due to sensor differences and calibration imperfections across devices \cite{das2015exploring, Neverova2016}. 
The source of such possible bias is avoided or reduced in the second case. Nevertheless, the performance of the developed authentication system could be highly affected if evaluated on a different device. 
The proposed BehavePassDB considers a hybrid approach, in the attempt to quantify a possible difference in the performance of a system developed in the first setting, which allows a simpler, large-scale data collection, and evaluated in both scenarios. We define the above-described case \textit{(i)} impostor data as \textit{random} forgeries, as the genuine and impostor data are acquired by two different devices, whereas case \textit{(ii)} is addressed as \textit{skilled impostor} forgeries, since the genuine and impostor data are related by the fact that the acquisition device is exactly the same and the impostor users were instructed to imitate the device owner, leading to a more challenging scenario (Fig. \ref{fig:captures}).  

\subsection{Contributions}
\label{subsec:Contributions}
\par In the present article, we aim to address the above mentioned aspects, contributing to the development of large-scale real-world of BBCA systems, as follows:
\begin{itemize}
\item We collect a multimodal behavioral biometrics database, BehavePassDB, involving 81 users, structured into separate acquisition sessions, in which the user is asked to carry out a series of tasks designed to mimic the most salient traits of mobile Human-Computer Interaction (HCI). The collected data also includes two dedicated sessions to be completed by another person (impostor) on the same exact device. 
In this way, we attempt to shed some light on the ability of the learned models to decorrelate user from device recognition in the learned data representation for background sensors. 
\item We consider several mobile behavioral biometric sources (touchscreen data, and background sensors such as accelerometer, gyroscope, magnetometer, lin. accelerometer, etc.). For each individual modality, we implement a popular Deep Learning (DL) architecture with the triplet loss function, and compare the biometric performance of individual modalities and their fusion at score level. 
\item The DL models are developed considering user data collected from different devices, and evaluated on user data collected in a similar scenario, and from the same exact device by an impostor user. Thus, we evaluate both \textit{random} and \textit{skilled} forgeries scenarios.
\item  We propose a standard experimental protocol publicly available to the research community in order to perform a fair comparison of novel approaches with the state of the art. This way we provide an easily reproducible framework. BehavePassDB\footnote{\texttt{\url{https://github.com/BiDAlab/MobileB2C_BehavePassDB/}}} and the experimental protocol followed in this paper are used to organize the MobileB2C\footnote{\texttt{\url{https://sites.google.com/view/mobileb2c/}}} competition at the International Joint Conference of Biometrics\footnote{\texttt{\url{http://www.ijcb2022.org/}}} (IJCB) 2022. This competition will be made an ongoing competition, so the current work will serve as a benchmark that researchers working in this field will be able to compare to.
\end{itemize}

\section{Related Work}
\label{sec:Related_Work}
\subsection{BBCA Systems}
\label{subsec:BBCA_Systems}
\par Given their suitability for continuous authentication on mobile devices, behavioral biometric traits have received the attention of the biometric research community since the launch of the first smartphone models. Over the years, different biometric systems have been developed and applied for BBCA, especially with the rise of DL. In the remainder of this section, some of the most recent and promising related studies up-to-date will be presented.
\par A mobile wide-ranging multimodal system based on behavioral biometrics was introduced by Deb \textit{et al.} \cite{Deb2019}. Their approach was based on a contrastive loss-based Siamese Long Short-Term Memory (LSTM) Recurrent Neural Network (RNN) architecture, in order to verify the users' identity in a passive way, i.e., without any explicit authentication task. 
8 different modalities were taken into account (keystroke, GPS location, accelerometer, gyroscope, magnetometer, linear accelerometer, gravity, and rotation sensors) and the individual scores were fused claiming results of 96.47\% of True Acceptance Rate (TAR) at a False Acceptance Rate (FAR) of 0.1\% considering 3-second time intervals for authentication. However, these results seem to be over-optimistic since they used a sampling frequency rate of 1Hz, i.e., 1 sample per second. They evaluated their system on a small self-collected dataset comprised of measurements from 30 smartphones for 37 subjects.
\par Following a similar method, Abuhamad \textit{et al.} proposed a mobile DL-based BBCA system only based on different sets of background sensors \cite{Abuhamad}. In this case, the fusion of modalites took place at data level, exploiting different RNN LSTM architectures with triplet loss. In 1-second time intervals, they obtained 0.41\% Equal Error Rate (EER) using three sensors (accelerometer, gyroscope, and magnetometer). The 84 participants in this study had to install a mobile application which transparently collected sensor information over time. Nonetheless, in both cases \cite{Deb2019, Abuhamad}, the time windows are extracted as long as the dedicated data acquisition app is running in the background, with loose restrictions in terms of sensor activity. A large amount of information is acquired per user, including many instances with little information content. 
Consequently, the average biometric distinctiveness of each single time window is allegedly less than in dense gesture-centered dedicated sessions, designed to mimic the most salient traits of mobile HCI. In light of this, it is difficult to assess how much of the authentication performance it to be attributed to the system identifying the device rather than the user.
\par Touchscreen data information was considered by Acien \textit{et al.} First, in \cite{2020_CDS_HCIsmart_Acien}, the HuMidb public database \cite{Acien2020b} was used to examine swiping gestures. A final EER of 19\% was achieved with a Siamese RNN by extracting 29 features. Then, the same authors adopted a LSTM RNN network for authentication based on keystroke dynamics in a free-text scenarios from the public Aalto database \cite{Palin_AaltoDBMobile19}, employing a variety of loss functions (softmax, contrastive, and triplet loss), different amount of enrolment data, length of the typed sequences, and device (physical vs touchscreen keyboard), achieving an EER of 9.2\% for touchscreen information while typing \cite{9539873}. 
\par In \cite{stragapede2022prl, stragapede2022iwbf}, several experiments are performed over HuMIdb, a separate RNN with triplet loss is implemented for each single modality with the weighted fusion of the different modalities is carried out at score level, leading to EER ranging from 4\% to 9\% depending on the modality combination in a 3-second interval.

\subsection{BBCA Databases}
\label{subsec:BBCA_Databases}
All machine learning-based systems thrive thanks to the availability of data, with no exception for the case of BBCA systems. The following aspects are important for a high-quality acquisition:
\begin{itemize}
    \item Involving a large number of subjects, maximizing the amount of data per user. This two aspects are often in conflict.
    \item Collecting data from several biometric sources (\textit{modalities}) in order to allow the development of multimodal systems, as such approach has proven to be beneficial in terms of robustness, immunity to noise, universality, and security, at a cost of increased complexity \cite{7503170}.
    %\cite{7503170, SINGH2019187, JAIN201680}.  
    \item Collecting data in an unconstrained scenario: this aspect is particularly sensitive with regard to behavioral biometrics for continuous authentication, given the ubiquity of mobile devices in the users' life and the diverse nature of such data. For instance, with respect to background sensors, an additional source of variability can be given by the user position or activity (sitting, standing, walking). Restricting the data acquisition scenario might affect the generality of the systems developed. Nevertheless, it is important to assess and avoid any possible learning bias due to users' position or activity.
    \item Number of acquisition devices: as well as the users' activity, also the usage of different acquisition devices can influence the ability of the systems to discriminate among human identities.
    Assessing this aspect rigorously is among the goals of the current study (see Sec. \ref{subsec:Device_Bias}).
    \item Public availability of the databases: assessing the performance of the different systems proposed in the literature is often a difficult task, given the different approaches, scopes, and the usage of self-collected non-public databases. Databases such as the Aalto database \cite{Palin_AaltoDBMobile19}, the UMDAA-02 \cite{Mahbub2016}, the HuMIdb \cite{Acien2020b}, etc., represent an important tool for the scientific community to compare approaches and advance the state of the art.
\end{itemize}
In Table I, we report some of the most important mobile behavioral biometric databases up to date.

\subsection{Device Bias}
\label{subsec:Device_Bias}
Das \textit{et al.} showed that under lab conditions a particular device could be identified by a response of its motion sensors to a given signal, developing a highly accurate fingerprinting mechanism that combines multiple motion sensors and makes use of (inaudible) audio stimulation to improve detection \cite{das2015exploring}. This happens due to imperfection in calibration of a sensor resulting in constant offsets and scaling coefficients (gains) of the output, that can be estimated by calculating integral statistics from the data. The authors analyzed techniques to mitigate such device fingerprinting either by calibrating the sensors to eliminate the signal anomalies, or by adding noise that obfuscates the anomalies. By acquiring measurements from a 30 different smartphones (5 models), they achieved a reduction of the accuracy by around 15\%-20\% in terms of average F-score by including additive and multiplicative noise to the raw data stream and spectral noise on the acquisition frequency. In light of this, it would be interesting to investigate how much of the authentication effectiveness should be attributed to the models extracting and recognizing features belonging to the device rather than the user. 
Following \cite{das2015exploring}, Neverova \textit{et al.} adopted such approach for large-scale study exploring temporal Deep Neural Networks (DNNs) for mobile biometric authentication based on accelerometer and gyroscope data \cite{Neverova2016}. The authors introduced low-level additive (offset) and multiplicative (gain) noise per training example to partially obfuscate the inter-device variations and ensure decorrelation of user identity from device signature in the learned data representation. They achieved a 93.3\% recognition accuracy by applying noise vectors obtained by drawing coefficients from a uniform distribution $\mu\sim\mathcal{U}_{12}[0.98, 1.02]$. Around 1500 subjects were involved, each one utilizing their own device (in all cases an LG Nexus 5). 

\begin{table*}
\centering
% \vspace*{-0.3\linewidth}
% \hspace*{-0.25\linewidth}
\label{tab:Mobile_BB_Databases}

\captionsetup{width=\textwidth}
\caption{Summary of Mobile Behavioral Biometric Databases$^1$. 
Acronyms: A = Accelerometer, Au = Audio, B = Bluetooth, Ba = Battery Level, C = Camera, °C = Temperature, CL = Call Logs, CT = Call Tower IDs, Gr = Gravity Sensor, Gy = Gyroscope, H = Handwriting, Hu = Humidity, K = Keystroke, L = Light, LA = Linear Accelerometer, Mi = Microphone, N = Network Logs, P = Pressure, Pr = Proximity, Sy = System Stats, T = Touchscreen, W = Wi-Fi.}
\begin{adjustbox}{width=\textwidth,center}
\normalsize
\begin{tabular}{l l c c c c c c}
\Xhline{2\arrayrulewidth}\textbf{Dataset} &\textbf{Year} &\textbf{Available$^2$} &\textbf{Unconstrained Env.$3$} &\textbf{\# of Users} &\textbf{\# of Devices} &\textbf{Data Modality} &\textbf{Impostor Case}\\
\Xhline{2\arrayrulewidth}
% MIT Reality Mining \cite{Eagle2006} & 2006 & \cmark & \cmark & 100 & 100 & \makecell{CL, B, CT, Ap} & Different Device \\
% \hline
% Saevanee \textit{et al} \cite{Saevanee2008} & 2008 & \xmark & \xmark & 10 & 1 & \makecell{T} & Same Device \\
% \hline
% Zahid \textit{et al.} \cite{Zahid2009} & 2009 & \xmark & \xmark & 25 & 13 & \makecell{T} & Not Considered \\	
% \hline
Rice LiveLab Dataset \cite{Shepard2011} & 2011 & \cmark & \cmark & 34 & 18 & \makecell{Ap, W} & Not Considered \\
\hline
Frank \textit{et al.} \cite{Frank2012} & 2012 & \xmark & \cmark & 41 & 4 & \makecell{T} & Not Considered \\
\hline
Serwadda \textit{et al.} \cite{Serwadda2013} & 2013 & \cmark & \xmark & 190 & 1 & \makecell{T} & Same Device \\
\hline
Zhang \textit{et al.} \cite{Zhang2015} & 2015 & \xmark & \xmark & 50 & 1 & \makecell{T, C} & Same Device \\
\hline
Feng \textit{et al.} \cite{Feng2014} & 2014 &  \xmark & \cmark & 123 & 3 & \makecell{T} & Not Considered \\
% \hline
% Xu \textit{et al.} \cite{Xu2014} & 2014 & \xmark & \xmark & 28 & 1 & \makecell{T, K, H} & Same Device \\
\hline
Saevanee \textit{et al.} \cite{Saevanee2015} & 2015 & \xmark & \cmark & 30 & $\sim$30 & \makecell{K, Ap, LP} & Different Device \\
% \hline
% Hoang \textit{et al.} \cite{Hoang2015} & 2015 & \xmark & \cmark & 34 & 1 & \makecell{A} & Same Device \\	
\hline
Neal \textit{et al.} \cite{Neal2015} & 2015 & \xmark & \cmark & 200 & $\sim$200 & \makecell{Ap, W, B} & Different Device \\
\hline
Wu \textit{et al.} \cite{Wu2015} & 2015 & \xmark & \cmark & 100 & $\sim$100 & \makecell{K, P, A, G} & Different Device \\
% \hline
% Nader \textit{et al.} \cite{Nader2015} & 2015 & \xmark & \xmark & 20 & 1 & \makecell{T} & Same Device \\
\hline
Murmuria \textit{et al.} \cite{Murmuria2015} & 2015 & \xmark & \cmark & 73 & 1 & \makecell{A, Gy, T, Ap, Ba} & Same Device \\
\hline
Sitova \textit{et al.} (HMOG) \cite{Sitova2015} & 2015 & \xmark & \xmark & 100 & 10 & \makecell{T, A} & Not Considered \\
% \hline
% Zaliva \textit{et al.} \cite{Zaliva2015} & 2015 & \xmark & \xmark & 14 & 1 & \makecell{T} & Same Device \\
\hline
Lu et Lio \cite{Lu2015} & 2015 & \xmark & \cmark & 60 & 3 & \makecell{T} & Not Considered \\
\hline
Coakley \textit{et al.} \cite{Coakley2016} & 2016 & \cmark & \xmark & 51 & 5 & \makecell{K, A, G} & Not Considered \\
% \hline
% Putri \textit{et al.} \cite{Putri2016} & 2016 & \cmark & \xmark & 29 & 1 & \makecell{T} & Same Device \\
\hline
Kumar \textit{et al.} \cite{Kumar2016} & 2016 & \xmark & \xmark & 28 & \textbf{-} & \makecell{T, K, A, G} & Not Considered \\
% \hline
% Shen \textit{et al.} \cite{Shen2015} & 2016 & \xmark & \xmark & 71 & 3 & \makecell{T} & Not Considered \\
\hline
Google Abacus Dataset \cite{Neverova2016} & 2016 & \xmark & \cmark & 1500 & $\sim$1500 & \makecell{C, T, A, Gy, M, W} & Different Device \\
% \hline
% Antal \textit{et al.} \cite{Antal2016} & 2016 & \xmark & \xmark & 71 & 8 & \makecell{T, A} & Not Considered \\
% \hline
% Nixon \textit{et al.} \cite{Nixon2016} & 2016 & \xmark & \cmark & 20 & $\sim$20 & \makecell{T, A, Gy} & Different Device \\
% \hline
% Buriro \textit{et al.} \cite{Buriro2016} & 2016 & \xmark & \xmark & 30 & 1 & \makecell{T, A, M, Gy} & Same Device \\
\hline
\makecell[l]{Mahbub \textit{et al.} \cite{Mahbub2016} (UMDAA-02)} & 2016 & \cmark & \cmark & 48 & \textbf{-} & \makecell{T, C, A, GPS, B,\\ W, L, P, °C, Pr} & Not Considered \\
\hline
Lee and Lee \cite{Lee2017} & 2017 & \xmark & \cmark & 35 & 2 & \makecell{A, Gy} & Not Considered \\		
\hline
Zhu \textit{et al.} \cite{Zhu2017} & 2017 & \xmark & \xmark & 20 & \textbf{-} & \makecell{A, Gy} & Not Considered \\
\hline
Al Kork \textit{et al.} \cite{AlKork2017} & 2017 & \xmark & \cmark & 50 & 2 & \makecell{A, Gy} & Not Considered \\
\hline
Li and Bours \cite{Li2018} & 2018 & \xmark & \cmark & 312 & $\sim$312 & \makecell{W, A} & Different Device \\
\hline
\makecell[l]{Amini \textit{et al.} \cite{Amini2018} (TargetAuth Dataset)} & 2018 & \xmark & \xmark & 47 & \textbf{-} & \makecell{A, Gy} & Not Considered \\
% \hline
% Cilia \textit{et al.} \cite{Cilia2018} & 2018 & \xmark & \xmark & 24 & 2 & \makecell{K} & Not Considered \\
\hline
Aalto University Dataset \cite{Palin_AaltoDBMobile19} & 2019 & \cmark & \cmark & $\sim$260k & \textbf{-} & \makecell{K} & Not Considered \\
\hline
Zhu \textit{et al.} \cite{Zhu2019} & 2019 & \xmark & Mixed & 1513 & 4 & \makecell{A, Gy, Gr} & Not Considered \\
% \hline
% Garbuz \textit{et al.} \cite{Garbuz2019} & 2019 & \xmark & \xmark & 36 & 1 & \makecell{T, A, Gy} & Same Device \\
% \hline
% Vajdi \textit{et al.} \cite{Vajdi2019} & 2019 & \cmark & \xmark & 93 & 2 & \makecell{A} & Not Considered \\		
\hline
Deb \textit{et al.} \cite{Deb2019} & 2019 & \xmark & \cmark & 37 & 30 & \makecell{K, GPS, A, Gy,\\ M, LA, Gr, Gy} & Not Considered \\
\hline
\makecell[l]{Acien \textit{et al.} \cite{Acien2020b} (HuMIdb)} & 2020 & \cmark & \cmark & 600 & 600 & \makecell{A, Gr, Gy, L, LA, M, O,\\ Pr, L, GPS, W, B, Mi} & Different Device \\
\hline
\makecell[l]{\textbf{Current Work (BehavePassDB)}} & \textbf{2022} & \textbf{\cmark} & \textbf{\cmark} & \textbf{81} & \textbf{81} & \makecell{\textbf{A, Gr, Gy, L, LA, M, O, P}\\ \textbf{Pr, L, GPS, W, B, °C, Ba, Hu}} & \makecell{\textbf{Different and Same}\\ \textbf{Device (Skilled)}} \\
\Xhline{2\arrayrulewidth}
\multicolumn{8}{l}{{\large $^1$ An extended version of the table is available at \texttt{\url{https://github.com/BiDAlab/MobileB2C_BehavePassDB/}}.}}\\
\multicolumn{8}{l}{{\large $^2$ Publicly Available.}}\\
\multicolumn{8}{l}{{\large $^3$ Unconstrained Environment, i.e., the subjects participating in the study did not receive instructions on how to perform the data collection process.}}\\
\end{tabular}
\end{adjustbox}
\end{table*}

\section{Description of BehavePassDB}
\label{sec:Description_of_the_BehavePassDB}
BehavePassDB includes data acquired during natural human-mobile interaction. The acquisition of the data was completed across four sessions, each of them separated by at least a 24-hour gap, in order to account for intra-subject variability. The participants were asked to install an Android application on their own smartphone and to complete eight tasks in an unsupervised scenario. The data collection process is designed to mimic the most salient scenarios of mobile HCI towards the idea of transparent CA, i.e., the user behavioral biometric traits get constantly verified throughout the device usage without completing any specific authentication process. Additionally, the acquisition sessions are structured to balance the amount of information captured and the collection effortlessness, to involve a large number of users. The tasks are as follows (Fig. \ref{fig:second_image}):

\begin{figure*}[t]
	\centering
% 	\hspace*{-0.25\linewidth}
    \includegraphics[width=\linewidth]{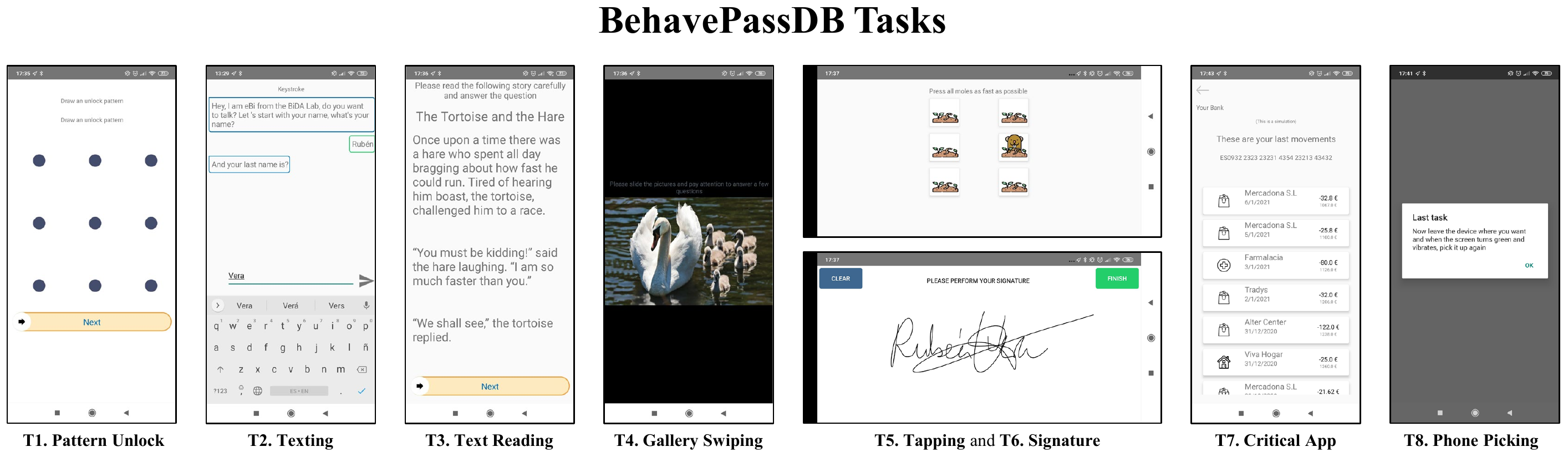}
    \captionsetup{width=\textwidth}
    \caption{Graphical representation of each of the 8 different tasks included in BehavePassDB data acquisition application.}
    \label{fig:second_image}
\end{figure*}

\begin{itemize}
    \item[\textbf{\textit{T1)}}] \textbf{Pattern Unlock}: the user performs a custom pattern unlock and a drag and drop gesture;
    \item[\textbf{\textit{T2)}}] \textbf{Texting}: the user is asked a series of different questions to be answered by typing. The questions are: name, last name, age, gender, two short free text questions, copying a number and a final free text question to be answered with at least 70 characters. Two different keyboards are deployed: in session 1 and 4, the keyboard presented to the user is their custom keyboard, whereas in session 2 and 3, we opt for a fixed keyboard. This choice is due to the fact that with the custom keyboard, it is typically possible to acquire only the timestamp and the value of the key pressed, from which a limited number of features can be extracted. The fixed keyboard, on the other hand, allows for the acquisition of the release time and the \textit{x} and \textit{y} coordinates of the touchscreen. Consequently, more features can be obtained. On the flip side, users might feel more natural when using their own preferred keyboard.
    \item[\textbf{\textit{T3)}}] \textbf{Text Reading}: the user is asked to read a text (swiping vertically) and then answer a question related to the text, followed by a drag and drop gesture.
    \item[\textbf{\textit{T4)}}] \textbf{Gallery Swiping}: the user is asked to swipe (horizontally) 4 pictures and then answer a question related to each of them, followed by a drag and drop gesture.
    \item[\textbf{\textit{T5)}}] \textbf{Tapping}: the user is asked to tap on in predetermined locations of the screen as fast as possible, followed by a drag and drop gesture.
    \item[\textbf{\textit{T6)}}] \textbf{Signature}: the user is asked to perform two handwritten signatures, one after the other.
    \item[\textbf{\textit{T7)}}] \textbf{Critical App}: this task is a simulation of a bank account app with different possible tasks inside, which are a mixture of the previous tasks.
    \item[\textbf{\textit{T8)}}] \textbf{Phone Picking}: the user is asked to leave the smartphone on a surface, wait for the screen to turn green, and then pick it up.
\end{itemize}
\par Simultaneously, data are acquired from 15 background sensors, i.e., touchscreen, accelerometer, linear accelerometer, gyroscope, magnetometer, ambient temperature, proximity, gravity, light, humidity, pressure, GPS, Wi-Fi, Bluetooth, and battery, collected as long as the device was able to provide such sensor information.
\par One of the most important novel aspect and novel contribution of BehavePassDB is its division into two subsets: a training set, containing 51 users, a validation set, containing 10 users, and an evaluation set, containing 20 users. In the first one, the acquisition scenario consists in each of the users using their own device, while the validation and evaluation sets also include two sessions of a different user on the owner's device. 
\par Each session consists in 8 tasks, common to all sessions. Sessions 1 and 2 serve as the enrolment sessions of the user data. In sessions 3 and 4 of the evaluation dataset there was also an additional task, which is a simulation of a critical app (bank account app) with a mix of the previous different tasks. The tasks are as follows (Fig. \ref{fig:second_image})
\par Regarding the age distribution, 8.86\% of the users were younger than 20 years old, 64.56\% are between 20 and 30 years old, 18.99\% between 30 and 50 years old, and the remaining 7.59\% are older than 50 years old. Regarding the gender, 46.84\% of the participants were males, 53.16\% females. The subjects were recruited from 8 countries (93.67\% European, 5.06\% American, 1.27\% Asian).

\section{System Description}
\label{sec:System_Description}
\subsection{Pre-processing and Feature Extraction}
\label{subsec:Pre-processing_and_Feature_Extraction}
\subsubsection{Background Sensor Data}
\label{subsec:Background_Sensor_Data}
The sensor data included in the current benchmark are acquired through the accelerometer, gravity sensor, gyroscope, linear accelerometer, and magnetometer.
The acquisition frequency is set to 200Hz, however, different devices can be equipped with different sensors whose specifications do not allow for such frequency value.
The time-series data are normalized through mean subtraction and division by the standard deviation, in the attempt to minimize the effect of noise and to cancel offset errors per axis per acquisition session. Following related studies \cite{Deb2019}, the Fast Fourier Transform (FFT) is computed from the raw \textit{x}, \textit{y}, \textit{z} values, and the first- and second-order derivatives are included as additional features. Preliminary experiments in fact showed the impact of the derivatives to be beneficial on the system performance and negligible in terms of computational burden. Finally, for each timestamp and sensor, the final output of the pre-processing operations consist in a 12-dimensional vector, as follows: 
\begin{center}
[\textit{x}, \textit{y}, \textit{z}, \textit{x'}, \textit{y'}, \textit{z'}, \textit{x''}, \textit{y''}, \textit{z''}, \textit{fft(x)}, \textit{fft(y)}, \textit{fft(z)}]
\end{center}

\subsubsection{Touch Data} 
\label{subsec:touch}
This benchmark is carried out considering the tasks of keystroke, text reading, gallery swiping, and tapping. The keystroke dynamics are assessed starting from data acquired during fixed questions with free answers, for instance related to the description of the last trip undertaken by the user, with a minimum answer length of 70 characters. The text acquired is in English or Spanish, depending on the preference of each subject. The entire sequence is acquired and analyzed, including the backspace key.
\par As explained in T2) (Sec. \ref{sec:Description_of_the_BehavePassDB}), the keystroke data are acquired with two different type of keyboards depending on the acquisition session. In the current work, we utilize the inter-press time and the normalized value of the ASCII code for all sessions. 
\par With regard to the tasks of text reading, gallery swiping, and tapping, the spatial \textit{x} and \textit{y} coordinates of the screen are used. Such data undergo a pre-processing process similar to the case of background sensor signals, including first- and second-order derivatives and the FFT, as reported below: 
\begin{center}
[\textit{x}, \textit{y}, \textit{x'}, \textit{y'}, \textit{x''}, \textit{y''}, \textit{fft(x)}, \textit{fft(y)}]
\end{center}
The \textit{x} and \textit{y} data are normalized as well to the height and width values of the screen to reject potential sources of bias across devices. 

\subsection{System Architecture}
\label{subsec:Learning_Architecture}
The proposed authentication system is based on an LSTM RNN, a DL network designed to exploit long-term dependencies in time-series data \cite{TOLOSANA2022108609, Deepsigntolosana}. The architecture implemented relies on two 64-unit layers with \textit{tan-h} activation functions. Additional steps include batch normalization, dropout  (with a rate of 0.5) between layers, and recurrent dropout (with a rate of 0.5) in each of the layers to limit the effect of overfitting. Preliminary tests led to the hyper-parameter configuration.
\par $M$-sample time windows are created by including consecutive data samples. The time windows are zero-padded if the obtained sequence is too short. In our system, a time window corresponds to the biometric information unit fed to the system for training and testing purposes. The size of the time window $M$ is modality-dependent ($M = 150$ for background sensors, $M = 100$ for all touch tasks except for keystroke, for which $M = 50$, and tap, for which $M = 20$). 
These values were chosen in order to obtain an adequate amount of information for touch gestures in a similar amount of time. 
\par For any $M$-sample time window, the corresponding DL model outputs a feature embedding, i.e., an $E$-dimensional array of real values that serves as a compact representation of the discriminative features hidden in the time-series data ($E = 64$ for all cases). The goal of each model is mapping time windows belonging to the same user to similar representations in the embedding space, whereas embeddings of different users should be as distant as possible from each other.

\subsection{Training Approach}
\label{subsec:Training_Approach}
\par For each modality, a separate unimodal network is trained, totaling 9, i.e., 4 touch tasks and 5 background sensors. Each of the background sensor models is developed by obtaining time windows evenly from each of the touch tasks. In fact, preliminary experiments proved this approach to be more effective, in comparison to developing a different network for every \textit{task-modality} combination. By doing so, the features learned by the background sensor networks are more general and robust.

\subsection{Triplet Loss Function} 
\label{subsec:Triplet_Loss_Function}
In the field of neural networks, the triplet loss is an extension of the constrastive loss function, allowing networks to learn simultaneously from positive and negative comparisons. The constrastive loss function, on the other hand, only allows one \textit{posivite-negative} comparison at once.
%\cite{JMLR:v10:weinberger09a}. 
A triplet is made of an ordered sequence of three independent time windows belonging to two different classes: the Anchor (A) and the Positive (P) are time windows extracted from different acquisition sessions of the same user, whereas Negative (N) is a time window from different user data. The triplet loss function is defined as follows: 
\begin{equation*}
\mathcal{L}_{TL} = max\{0, d^2(\mathbf{v}_{A}, \mathbf{v}_{P}) - d^2(\mathbf{v}_{A}, \mathbf{v}_{N}) + \alpha \}
\end{equation*}
where $\alpha$ is the margin between positive and negative pairs and $d$ is the Euclidean distance between \textit{anchor}-\textit{positive} $(\mathbf{v}_{A}$-$\mathbf{v}_{P})$ pairs and \textit{anchor}-\textit{negative} $(\mathbf{v}_{A}$-$\mathbf{v}_{N})$ pairs ($\alpha = 1.5$). The triplet loss function is employed to minimize the distance between embedding vectors from the same class $(d^2(\mathbf{v}_{A}, \mathbf{v}_{P}))$, and to maximize it for different class embeddings $(d^2(\mathbf{v}_{A}, \mathbf{v}_{N}))$ in a single step.

\subsection{Fusion of Modalities}
\label{subsec:Fusion_of_Modalities}
In the area of biometrics, a variety of fusion methods have been proposed in the literature.
%\cite{ROSS20032115}. 
In the current work, we adopt the fusion at score level, i.e., the fusion is achieved through a linear combination of the scores. In particular, the scores consist in the Euclidean distances between embeddings computed starting from simultaneous time windows pertaining to different modalities. In this way, the authentication system benefits from modularity.
In fact, in the proposed authentication system, the independent models can be included at different times or easily be replaced, if their output embeddings have the same size. Consequently, the separate models can be improved individually, leaving margin for improvement. 
During any of the touch tasks, six modalities at most are combined: the touch information from each task (keystroke, text reading, gallery swiping, tapping), and the five background sensors (accelerometer, gravity sensor, gyroscope, linear accelerometer, magnetometer), yielding 63 different fusion combinations.

\section{Experimental Protocol}
\label{sec:Experimental_Protocol}
\par BehavePassDB is divided into three subsets: \textit{(i)} the training set, including 51 users, in which the data have been collected with every user using their own device, \textit{(ii)} the validation set, including 10 users, and \textit{(iii)} the evaluation set, containing 20 users. The last two sets include the same acquisition scenario considered in the case of the training set (each user using their own device), but also the scenario of an impostor user using the same device as the owner in the last two of the four acquisition sessions. Consequently, as the training set only contains the random impostor cases, the models are optimized to compare random forgeries rather than skilled ones.
\par The assessment of the performance of the network is based on the comparison of embeddings belonging to each of the users with their own and with other users' embeddings. In all phases (training, validation, and evaluation), the first two sessions of each user are considered for enrolment, whereas the remaining two for verification. 
\par With regard to the hyper-parameters in training, each of the models is trained for 150 epochs, the batch size is 512, the learning rate is 0.05, the Adam optimizer is employed with $\beta_{1} = 0.9$, $\beta_{2} = 0.999$ and $\epsilon = 10^{-8}$. \texttt{Keras-Tensorflow} is used to develop the models. 
\par The training of the networks takes place by randomly withdrawing initial time instants from the whole duration of the time sequence of each of the acquisition sessions, always guaranteeing, if possible, full $M$-sample time windows and avoiding zero-padding. The optimization of the model parameters has taken place in a preliminary phase of the experimental work presented using the development and validation datasets.
\par For the purpose of validation and evaluation, data from two sessions at a time are considered to compute a final distance score value. Up to 50 overlapping time windows per session (for the case of keystroke the overlap $\Delta$ = 20, for the cases of text reading, gallery swiping and tap $\Delta$ = 10, and for the case of all other background sensors $\Delta$ = 50) are considered. For each of the time windows, the corresponding output embedding is computed. The final distance score is  computed by averaging the Euclidean distance values obtained from all possible combination pairs of output embeddings across the two considered sessions. This approach leads to a single distance value for each session-to-session comparison. The obtained distance value is included in the genuine score distribution if the sessions concern the same user, and in the impostor score distribution otherwise. To achieve a multimodal system by fusion at score level, we sum the distance values obtained from each of the different modalities considered and pertaining to the same session.
For each of the subjects in the validation and evaluation dataset, we obtain the score distributions as follows:
\begin{itemize}
    \item genuine distribution: 2 values per user in the set, obtained comparing the 2 genuine verification sessions with the user's enrolment sessions;
    \item random impostor distribution: 2 values per user in the set, obtained comparing 2 verification sessions of a different user with the user's enrolment sessions;
    \item skilled impostor distribution: 2 values per user in the set, obtained comparing 2 skilled impostor sessions acquired on the same device with the user's enrolment sessions.
\end{itemize}
The performance scores described in Sec. \ref{sec:Experimental_Results} are obtained considering the overall distributions as described above. In the current article, the experimental protocol and database adopted are the same as the ones used in the recently proposed Mobile Behavioral Biometrics Competition (MobileB2C) at the International Conference of Joint Biometrics (IJCB) 2022, which will be made ongoing. Consequently, the current article serves as a first, thorough benchmark.

\section{Experimental Results}
\label{sec:Experimental_Results}
{\subsection{\color{orange}Random Impostor Case}}
\label{subsec:Random_Impostor_Case}

\subsubsection{Individual Modalities}
\label{subsubsec:Individual_Modalities_R}
Table \ref{tab:unimodal_moda_random} shows the results obtained on the evaluation set (the validation set scores are in brackets) considering each modality individually in the random impostor scenario. The metric chosen to evaluate the system performance is the Area Under the Curve (AUC) of the Receiving Operating Characteristic (ROC). The ROC curve is drawn by considering the False Positive Rate (FPR) on the \textit{x}-axis, and the True Positive Rate (TPR) on the \textit{y}-axis. The former indicates the rate of non-detections of impostor scores out of all acceptances, whereas the latter indicates those correctly detected as impostor scores out of all rejections. The AUC quantifies the overall ability of the system to discriminate between two classes. Random guessing has an AUC of 50\%, whereas a perfect system has an AUC of 100\%. Each of the rows of Table \ref{tab:unimodal_moda_random} presents the results pertaining to the touchscreen data of single task, and the corresponding background sensor data acquired simultaneously in the following columns. 
In terms of unimodal touchscreen information performance, the most effective task is text reading, achieving 73.22\% AUC. Then, the gallery swiping and the tapping tasks achieve an AUC performance of respectively 63.58\% and 64.56\%. Finally, an AUC score of 57.48\% is obtained during the keystroke task. Such values individually are far from satisfactory. However, such results show that in this experimental setup it is possible to extract more discriminative information from the text reading dynamics than in the case of the other tasks. The last row of Table \ref{tab:unimodal_moda_random} shows the AUC scores achieved by each background sensor averaged over the tasks. The magnetometer and the linear accelerometer consistently proves to be the best performing sensors (respectively 73.03\% and 75.43\% AUC), while the accelerometer, gravity sensor, and gyroscope do not reach similar results (respectively 61.80\%, 60.61\% and 62.86\% AUC). 

\subsubsection{Fusion of Modalities}
\label{subsubsec:Fusion_of_Modalities_R}
\par In Table \ref{tab:multimodal_moda_random}, the best subsets originated from the fusion of modalities are included. In any case, the improvement in the performance of the system due to the fusion of modalities is significant. The individual modalities altogether achieve an average 65.84\% AUC, whereas the average AUC of the best modality combinations is 82.47\%, i.e., a relative improvement of 25\% AUC. 
The best performance is achieved for the task of keystroke with a fusion of touch, gyroscope, lin. accelerometer, and magnetometer, reaching 87.20\% AUC. In this case, the AUC produced by the fusion of modalities is around 30\% higher in absolute terms compare to the one achieved with the touch data only. The other modalities reach a level of performance around 80\% AUC.

\begin{table*}[pt!]
% \vspace*{-0.27\textwidth}
\captionsetup{width=\textwidth}
\caption{Results in terms AUC (\%) of the different \textbf{individual modalities for each task in the {{\color{orange}random impostor scenario}}}. In brackets the results obtained on the validation set are displayed. The best results are highlighted in bold.}
\centering
% \small
\label{tab:unimodal_moda_random}
\begin{adjustbox}{width=\textwidth,center}
\begin{tabular}{l l|l|l|l|l|l|}
% \multicolumn{7}{ c }{\textbf{Individual Modalities}}\tabularnewline
\multicolumn{1}{ l }{Task} & \multicolumn{1}{ c }{Touchscreen} & \multicolumn{1}{l}{Accelerometer} & \multicolumn{1}{l}{Gravity Sensor} & \multicolumn{1}{l}{Gyroscope} & \multicolumn{1}{l}{Lin. Accelerometer} & \multicolumn{1}{l}{Magnetometer} \tabularnewline
\Xhline{2\arrayrulewidth}
\textbf{Keystroke} & \multicolumn{1}{c}{57.48 (72.19)} & \multicolumn{1}{c}{66.23 (68.94)} & \multicolumn{1}{c}{63.84 (57.56)} & \multicolumn{1}{c}{66.47 (65.75)} & \multicolumn{1}{c}{79.25 (78.13)} & \multicolumn{1}{c}{\textbf{81.55} (65.50)} \tabularnewline
\hline
\textbf{Text Reading} & \multicolumn{1}{c}{\textbf{73.22} (67.00)} & \multicolumn{1}{c}{58.61 (58.99)} & \multicolumn{1}{c}{57.28 (50.13)} & \multicolumn{1}{c}{59.66 (55.44)} & \multicolumn{1}{c}{64.72 (67.63)} & \multicolumn{1}{c}{\textbf{72.39} (64.87)} \tabularnewline
\hline
\textbf{Gallery Swiping} & \multicolumn{1}{c}{63.58 (57.88)}& \multicolumn{1}{c}{62.08 (63.00)} & \multicolumn{1}{c}{60.47 (60.00)} & \multicolumn{1}{c}{60.75 (62.00)} & \multicolumn{1}{c}{\textbf{77.50} (79.36)} & \multicolumn{1}{c}{75.20 (77.31)} \tabularnewline
\hline
\textbf{Tapping} & \multicolumn{1}{c}{64.56 (67.38)} & \multicolumn{1}{c}{60.27 (55.31)} & \multicolumn{1}{c}{60.83 (56.69)} & \multicolumn{1}{c}{64.56 (58.75)} & \multicolumn{1}{c}{70.66 (71.88)} & \multicolumn{1}{c}{\textbf{72.58} (73.25)} \tabularnewline
\hline
\multicolumn{2}{l}{\textbf{Average of Background Sensors}}& \multicolumn{1}{c}{61.80 (61.56)} & \multicolumn{1}{c}{60.61 (56.10)} & \multicolumn{1}{c}{62.86 (60.49)} & \multicolumn{1}{c}{73.03 (74.25)} & \multicolumn{1}{c}{\textbf{75.43} (70.26)} \tabularnewline
\end{tabular}
\end{adjustbox}
% \end{table*}
% \vspace*{-0.2\textwidth}
% \begin{table*}[h!]
\captionsetup{width=\textwidth}
\caption{Results in terms of AUC (\%) of the best subsets originated from \textbf{the fusion of the different individual modalities for each task in the {\color{orange}random impostor scenario}}. In brackets the results obtained on the validation set are displayed. The best result is highlighted in bold.}
\centering
\normalsize
\label{tab:multimodal_moda_random}
\begin{tabular}{l|l|l|}
% \multicolumn{3}{ c }{\textbf{Fusion of Modalities}}\tabularnewline
\multicolumn{1}{l}{Task} & 
\multicolumn{1}{c}{AUC (\%)} & \multicolumn{1}{c}{Best Modality Subset}
\tabularnewline
% \Xhline{2\arrayrulewidth}
\multicolumn{1}{l}{\textbf{Keystroke}} & 
\multicolumn{1}{c}{\textbf{87.20} (83.56)} 
& \multicolumn{1}{c}{\textbf{K, Gy, L, M} (K, A, Gy, L)} 
\tabularnewline
\hline
\multicolumn{1}{l}{\textbf{Text Reading}} &
\multicolumn{1}{c}{81.31 (78.75)} &
\multicolumn{1}{c}{TR, Gr, M (TR, A, L, M)}
\tabularnewline
\hline
\multicolumn{1}{l}{\textbf{Gallery Swiping}} & 
\multicolumn{1}{c}{81.58 (84.56)} & \multicolumn{1}{c}{Gr, L, M (GS, L, M)} 
\tabularnewline
\hline
\multicolumn{1}{l}{\textbf{Tap}} &
\multicolumn{1}{c}{79.80 (81.50)} &
\multicolumn{1}{c}{Gr, Gy, L, M (T, Gy, L, M)} 
\tabularnewline
\hline
\multicolumn{3}{p{0.55\linewidth}}{Acronyms of Tasks: K = Keystroke, TR = Text Reading, GS = Gallery Swiping, T = Tap. Acronyms of Background Sensors: A = Accelerometer, Gr = Gravity Sensor, Gy = Gyroscope, L = Linear Accelerometer, M = Magnetometer.}
\end{tabular}
% \end{table*}
% \vspace*{-0.2\textwidth}
% \begin{table*}[h!]
% \captionsetup{width=\textwidth}
\caption{Results in terms of AUC (\%) of the different \textbf{individual modalities for each task in the {\color{red}skilled impostor scenario}}. In brackets the results obtained on the validation set are displayed. The best results are highlighted in bold.}
\centering
\label{tab:unimodal_moda_skilled}
\begin{adjustbox}{width=\textwidth,center}
\begin{tabular}{l l|l|l|l|l|l|}
% \multicolumn{7}{ c }{\textbf{Individual Modalities}}\tabularnewline
\multicolumn{1}{ l }{Task} & \multicolumn{1}{ c }{Touchscreen} & \multicolumn{1}{l}{Accelerometer} & \multicolumn{1}{l}{Gravity Sensor} & \multicolumn{1}{l}{Gyroscope} & \multicolumn{1}{l}{Lin. Accelerometer} & \multicolumn{1}{l}{Magnetometer} \tabularnewline
\Xhline{2\arrayrulewidth}
\textbf{Keystroke} & \multicolumn{1}{c}{56.18 (60.63)} & \multicolumn{1}{c}{56.22 (65.88)} & \multicolumn{1}{c}{59.43 (58.06)} & \multicolumn{1}{c}{58.89 (60.75)} & \multicolumn{1}{c}{\textbf{67.28} (59.75)} & \multicolumn{1}{c}{60.27 (54.56)} \tabularnewline
\hline
\textbf{Text Reading} & \multicolumn{1}{c}{\textbf{66.05} (69.12)} & \multicolumn{1}{c}{52.92 (54.56)} &
\multicolumn{1}{c}{\textbf{56.31} (59.75)} & 
\multicolumn{1}{c}{50.78 (51.37)} & 
\multicolumn{1}{c}{53.33 (54.50)} & 
\multicolumn{1}{c}{50.67 (55.13)} \tabularnewline
\hline
\textbf{Gallery Swiping} & \multicolumn{1}{c}{62.22 (67.25)}& 
\multicolumn{1}{c}{51.45 (55.56)} & 
\multicolumn{1}{c}{53.77 (61.25)} & 
\multicolumn{1}{c}{60.53 (52.62)} & 
\multicolumn{1}{c}{\textbf{63.73} (54.06)} & 
\multicolumn{1}{c}{57.36 (57.00)} \tabularnewline
\hline
\textbf{Tapping} & \multicolumn{1}{c}{59.11 (51.13)} & \multicolumn{1}{c}{\textbf{60.98} (50.50)} & \multicolumn{1}{c}{55.88 (58.81)} & \multicolumn{1}{c}{\textbf{60.98} (64.19)} & \multicolumn{1}{c}{55.88 (68.38)} & \multicolumn{1}{c}{54.08 (67.06)} \tabularnewline
\hline
\multicolumn{2}{l}{\textbf{Average of Background Sensors}}& \multicolumn{1}{c}{55.39 (58.22)} & \multicolumn{1}{c}{56.35 (59.47)} & \multicolumn{1}{c}{57.80 (57.23)} & \multicolumn{1}{c}{\textbf{60.06} (59.17)} & \multicolumn{1}{c}{55.60 (58.44)} \tabularnewline
\end{tabular}
\end{adjustbox}
% \end{table*}
% \vspace*{-0.5\textwidth}
% \begin{table*}[b!]
\captionsetup{width=\textwidth}
\caption{Results in terms of AUC (\%) of the best subsets originated from \textbf{the fusion of the different individual modalities for each task in the {\color{red}skilled impostor scenario}}. In brackets the results obtained on the validation set are displayed. The best result is highlighted in bold.}
\centering
\normalsize
\label{tab:multimodal_moda_skilled}
\begin{tabular}{l|l|l|}
% \multicolumn{3}{ c }{\textbf{Fusion of Modalities}}\tabularnewline
\multicolumn{1}{l}{Task} & 
\multicolumn{1}{c}{AUC (\%)} & \multicolumn{1}{c}{Best Modality Subset}
\tabularnewline
% \Xhline{2\arrayrulewidth}
\multicolumn{1}{l}{\textbf{Keystroke}} &
\multicolumn{1}{c}{\textbf{68.72} (70.50)} & \multicolumn{1}{c}{\textbf{K, A, Gr, Gy, L, M} (K, A, Gr, Gy, L)}
\tabularnewline
\hline
\multicolumn{1}{l}{\textbf{Text Reading}} &
\multicolumn{1}{c}{66.73 (69.12)} &
\multicolumn{1}{c}{TR, Gr (TR, A, L, M)}
\tabularnewline
\hline
\multicolumn{1}{l}{\textbf{Gallery Swiping}} & 
\multicolumn{1}{c}{67.52 (67.25)} & \multicolumn{1}{c}{GS, Gy, L (GS)} 
\tabularnewline
\hline
\multicolumn{1}{l}{\textbf{Tap}} &
\multicolumn{1}{c}{61.92 (70.62)} &
\multicolumn{1}{c}{T, A, L (Gy, L)} 
\tabularnewline
\hline
\multicolumn{3}{p{0.6\linewidth}}{ Acronyms of Tasks: K = Keystroke, TR = Text Reading, GS = Gallery Swiping, T = Tap. Acronyms of Background Sensors: A = Accelerometer, Gr = Gravity Sensor, Gy = Gyroscope, L = Linear Accelerometer, M = Magnetometer.}
\end{tabular}
\end{table*}

\begin{table}[t!]
\captionsetup{width=0.45\textwidth}
\caption{Results in terms of \textit{p}-value of the best subsets originated from \textbf{the fusion of the different individual modalities for each task in the {\color{orange}random} and {\color{red}skilled} impostor scenarios}.}
\centering
\label{tab:wilcox}
\begin{tabular}{cl|l|l|}
% \multicolumn{3}{ c }{\textbf{Fusion of Modalities}}\tabularnewline
\multicolumn{1}{c}{Case} & 
\multicolumn{1}{c}{Task} & 
\multicolumn{1}{c}{Best Modality Subset} &
\multicolumn{1}{c}{\textit{p}-value}
\tabularnewline
\Xhline{2\arrayrulewidth}
\multirow{4}{*}{\rotatebox[origin=c]{270}{\color{orange}Random}} &
\multicolumn{1}{l}{\textbf{Keystroke}} &
\multicolumn{1}{c}{K, Gy, L, M} &
\multicolumn{1}{c}{4.46E-16}
\tabularnewline
\cline{2-4}
&
\multicolumn{1}{l}{\textbf{Text Reading}} &
\multicolumn{1}{c}{TR, Gr, M} &
\multicolumn{1}{c}{7.98E-12} 
\tabularnewline
\cline{2-4}
&
\multicolumn{1}{l}{\textbf{Gallery Swiping}} & 
\multicolumn{1}{c}{Gr, L, M} &
\multicolumn{1}{c}{5.31E-12} 
\tabularnewline
\cline{2-4}
&
\multicolumn{1}{l}{\textbf{Tap}} &
\multicolumn{1}{c}{Gr, Gy, L, M} & 
\multicolumn{1}{c}{7.62E-11}
\tabularnewline
\Xhline{2\arrayrulewidth}
\multirow{4}{*}{\rotatebox[origin=c]{270}{\color{red}Skilled}} &
\multicolumn{1}{l}{\textbf{Keystroke}} &
\multicolumn{1}{c}{K, A, Gr, Gy, L, M} & 
\multicolumn{1}{c}{1.90E-15} 
\tabularnewline
\cline{2-4}
&
\multicolumn{1}{l}{\textbf{Text Reading}} &
\multicolumn{1}{c}{TR, Gr} &
\multicolumn{1}{c}{1.20E-7}
\tabularnewline
\cline{2-4}
&
\multicolumn{1}{l}{\textbf{Gallery Swiping}} & 
\multicolumn{1}{c}{GS, Gy, L} &
\multicolumn{1}{c}{7.13E-9} 
\tabularnewline
\cline{2-4}
&
\multicolumn{1}{l}{\textbf{Tap}} &
\multicolumn{1}{c}{T, A, L} & 
\multicolumn{1}{c}{4.73E-6}
\tabularnewline
\multicolumn{4}{p{0.9\linewidth}}{Acronyms of Tasks: K = Keystroke, TR = Text Reading, GS = Gallery Swiping, T = Tap. Acronyms of Background Sensors: A = Accelerometer, Gr = Gravity Sensor, Gy = Gyroscope, L = Linear Accelerometer, M = Magnetometer.}
\end{tabular}
\end{table}

\subsection{{\color{red}Skilled Impostor Case}}
\label{subsec:Skilled_Impostor_Case}

\subsubsection{Individual Modalities}
\label{subsubsec:Individual_Modalities_S}
Table \ref{tab:unimodal_moda_skilled} shows the results obtained on the evaluation set considering each modality individually in the skilled impostor scenario. The table is structured as its counterpart for the random impostor scenario. Once again the most discriminative touch modality is text reading, achieving 66.05\% AUC. Then, gallery swiping produces a score 62.22\% AUC. Finally, an AUC score below 60\% is obtained during the tasks of keystroke and tapping. Such values individually are slightly lower than in the case of the random impostor scenario. The last row of Table \ref{tab:unimodal_moda_skilled} shows the AUC scores achieved by each background sensor averaged over the tasks. In this case linear accelerometer proves to be the best performing sensor (60.06\% AUC), followed by gyroscope (57.80\% AUC), gravity sensor (56.35\% AUC), magnetometer (55.60\% AUC), and accelerometer (55.39\% AUC). %The injection of noise, once again, does not show signs of improving the behavior of the system.

\subsubsection{Fusion of Modalities}
\label{subsubsec:Fusion_of_Modalities_S}
Table \ref{tab:multimodal_moda_skilled} shows the best subsets originated from the fusion of modalities for the skilled impostor scenario. The fusion of the different modalities is generally beneficial for the system performance in the skilled impostor case as well, even though the overall performance is not as good compared to the random impostor scenario. This is in fact a much more challenging scenario in which the impostor user can imitate the behavior of the genuine user and also they both use exactly the same device. The individual modalities achieve an average 58.75\% AUC, whereras the average AUC of the best modality combinations is 66.22\% AUC, i.e., a relative improvement of 12.71\% AUC. 
The best performance is achieved for the task of keystroke with a fusion of all modalities, reaching 68.72\% AUC. In this case, the AUC produced by the fusion of modalities is around 10\% higher in absolute terms compared to the one achieved with the touch data only. The performance of the remaining modalities is comparable, apart from the task of tapping (60.98\% of AUC), which does not improve very much.

\subsection{Wilcoxon Rank-Sum Test}
\label{subsec:wilcoxon}
In order to assess the statistical significance of the results obtained, the Wilcoxon rank-sum test\footnote{\texttt{\url{https://docs.scipy.org/doc/scipy/reference/generated/scipy.stats.ranksums.html}}} (also known as or Wilcoxon–Mann–Whitney test) is carried out. It consists in a nonparametric test of the null hypothesis that, for randomly selected values \textit{X} and \textit{Y} from two populations, the probability of \textit{X} being greater than \textit{Y} is equal to the probability of \textit{Y} being greater than \textit{X}, indicating that it is impossible to determine that the selected values belong to different populations. In this case, the two populations are represented by the distance values respectively obtained from all the genuine and impostor embedding pairwise comparisons considered in the adopted experimental protocol. The alternative hypothesis is that values in one population (impostor distance value distribution) are more likely to be larger than the values in the other (genuine distance value distribution). The Wilcoxon rank-sum test yields the p-value, which represents the probability that the null hypothesis is true. Typically, a \textit{p}-value less than 0.05 is considered strong evidence against the null hypothesis, as there is less than a 5\% probability that such hypothesis is correct.
\par A separate Wilcoxon rank-sum test is carried out for each of the tasks and each of the impostor scenarios (random and skilled), considering the best modality subsets displayed in Table \ref{tab:multimodal_moda_random} and Table \ref{tab:multimodal_moda_skilled}), totaling the eight cases presented in Table \ref{tab:wilcox}. In all cases, the \textit{p}-values obtained span in a range between $10^{-6}$ and $10^{-16}$ depending on the specific task and impostor case, thus rejecting the null hypothesis, showing that it is possible to distinguish genuine and impostor data.

% \newpage
\subsection{Discussion}
\label{subsec:discussion}
Interesting conclusions can be drawn by comparing the results presented in the tables above. 
\par The beneficial effects of the fusion of the different modalities is consistent, and it is greater in the cases where the fused modalities achieve already good results individually. Consequently, developing unimodal models able to perform well is a key aspect in order to improve the system performance.
\par
\begin{figure*}[h]
\centering
\phantomsection\label{fig:ROCcurves}
\minipage{0.5\textwidth}
\adjincludegraphics[width=\linewidth,trim={0 0 {.11\width} {.11\width}}]{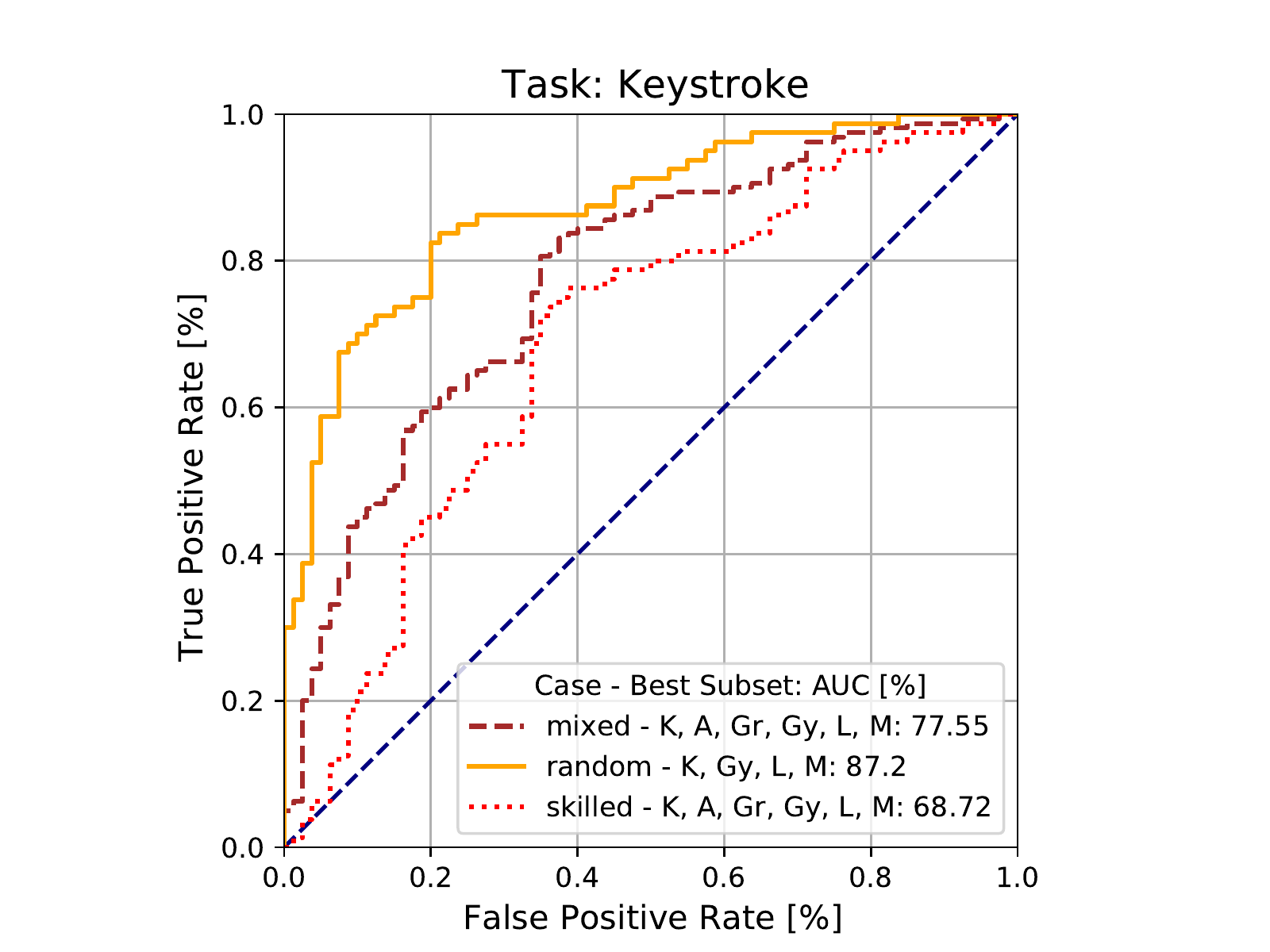}
\endminipage
\minipage{0.5\textwidth}
\adjincludegraphics[width=\linewidth,trim={0 0 {.11\width} {.11\width}}]{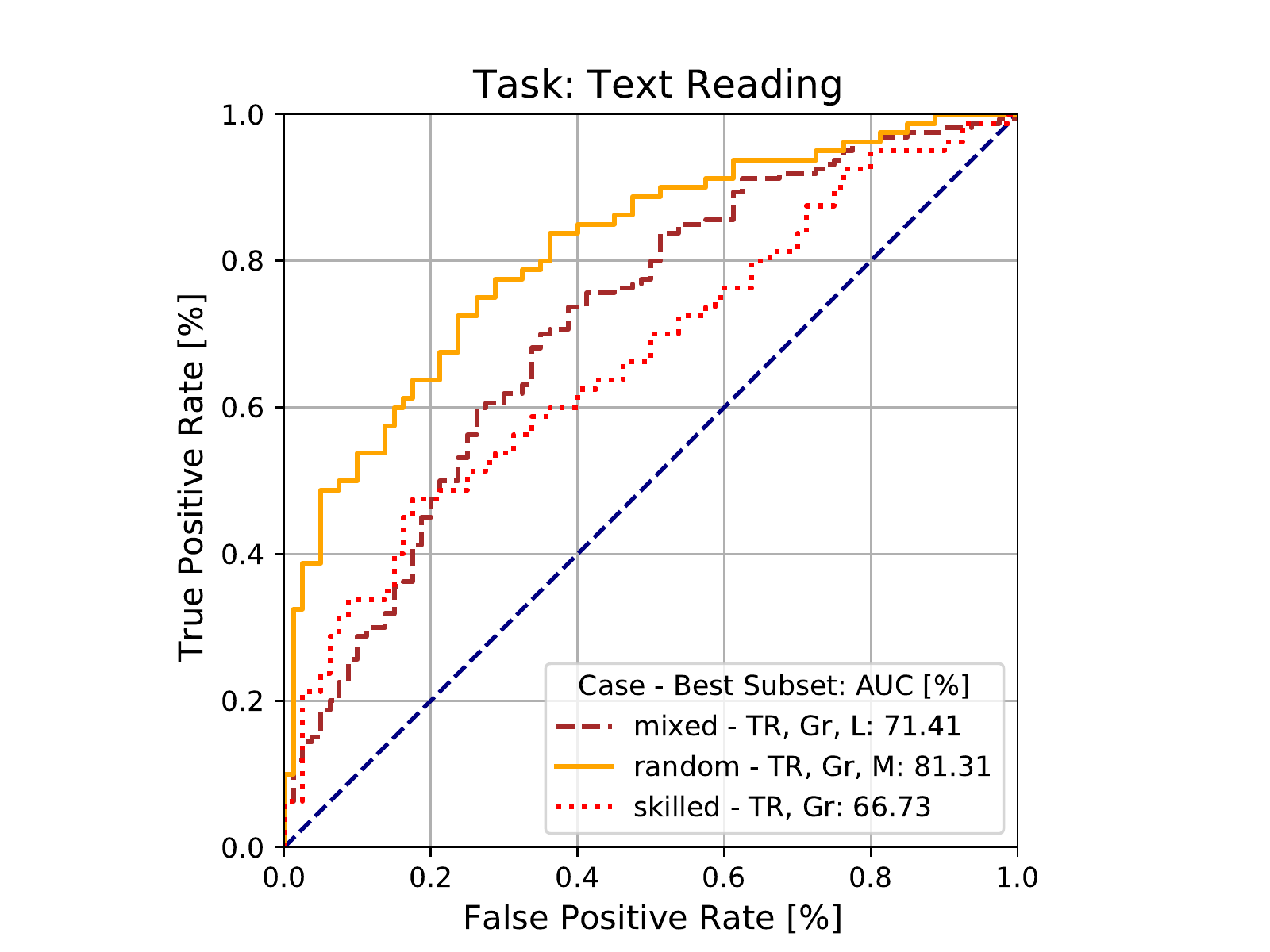}
\endminipage
\vspace*{0.1cm}
\end{figure*}

 With regard to the quantification of the device bias in the learned representation, we can compare the results shown in Table \ref{tab:unimodal_moda_random} with those of Table \ref{tab:unimodal_moda_skilled}. In the last row of the two tables it is possible to see the values averaged over the different tasks. All background sensors achieve lower performance in the more challenging skilled scenario. Such trend is graphically represented also in Fig. \hyperref[fig:ROCcurves]{3}, which shows the ROC curves for the four tasks considered. Each of the graph is related to a different task and shows the ROC curve for the best fusion subset considering both the described impostor scenarios. Additionally, it is possible to see the \textit{mixed scenario}, in which the impostor distribution is obtained including both previously described impostor cases.
In particular, the greatest impact is on the magnetometer and lin. accelerometer (respectively 20\% and 13\% lower in absolute terms), whereas the other modalities only undergo a 5\% AUC reduction. This trend might be due to the fact that these two sensors have a greater device bias impact in the performance, reflecting more the device fingerprint compared to the others. However, it should be pointed out that this is a more challenging case as the impostor user is next to the genuine one and could imitate the dynamics of the genuine user in a better way. Moreover, the training set of the models only contains the random impostor cases, consequently they are optimized to compare random forgeries rather than skilled ones. In order to further improve the skilled scenario, it would be needed to also have examples of skilled forgeries during the training process, but such data collection on a larger scale can be difficult to achieve, while guaranteeing the same realistic conditions as the evaluation dataset.
On the flip side, such bias could be exploited to implement a transparent security technology as a second factor in a 2-factor authentication (2FA) process, as could be a remote
security-wise critical mobile application, as every user would be using their own mobile device. Such trend also characterizes the touch tasks, although the AUC reduction is slightly less.
\par In comparison with recent related studies in the field \cite{Abuhamad, Deb2019} (see Sec. \ref{sec:Related_Work}), although adopting a similar biometric system, the authentication results achieved in this article are not as high. This trend could be due to the fact that BehavePassDB is a very challenging database, especially for the dedicated skilled impostor case, as it is designed to to investigate the feasibility of BBCA systems in different real-world scenarios. In the field of mobile behavioral biometrics, it is in fact often difficult to reach a global and significant conclusion from the comparison of different systems, given the different approaches, scopes, metrics, and the usage of self-collected non-public databases. Therefore, we aim to provide a useful tool to the biometric community to advance towards different application use cases and impostor scenarios.

\begin{figure*}[h]
\centering
\minipage{0.5\textwidth}
\adjincludegraphics[width=\linewidth,trim={0 0 {.11\width} {.11\width}}]{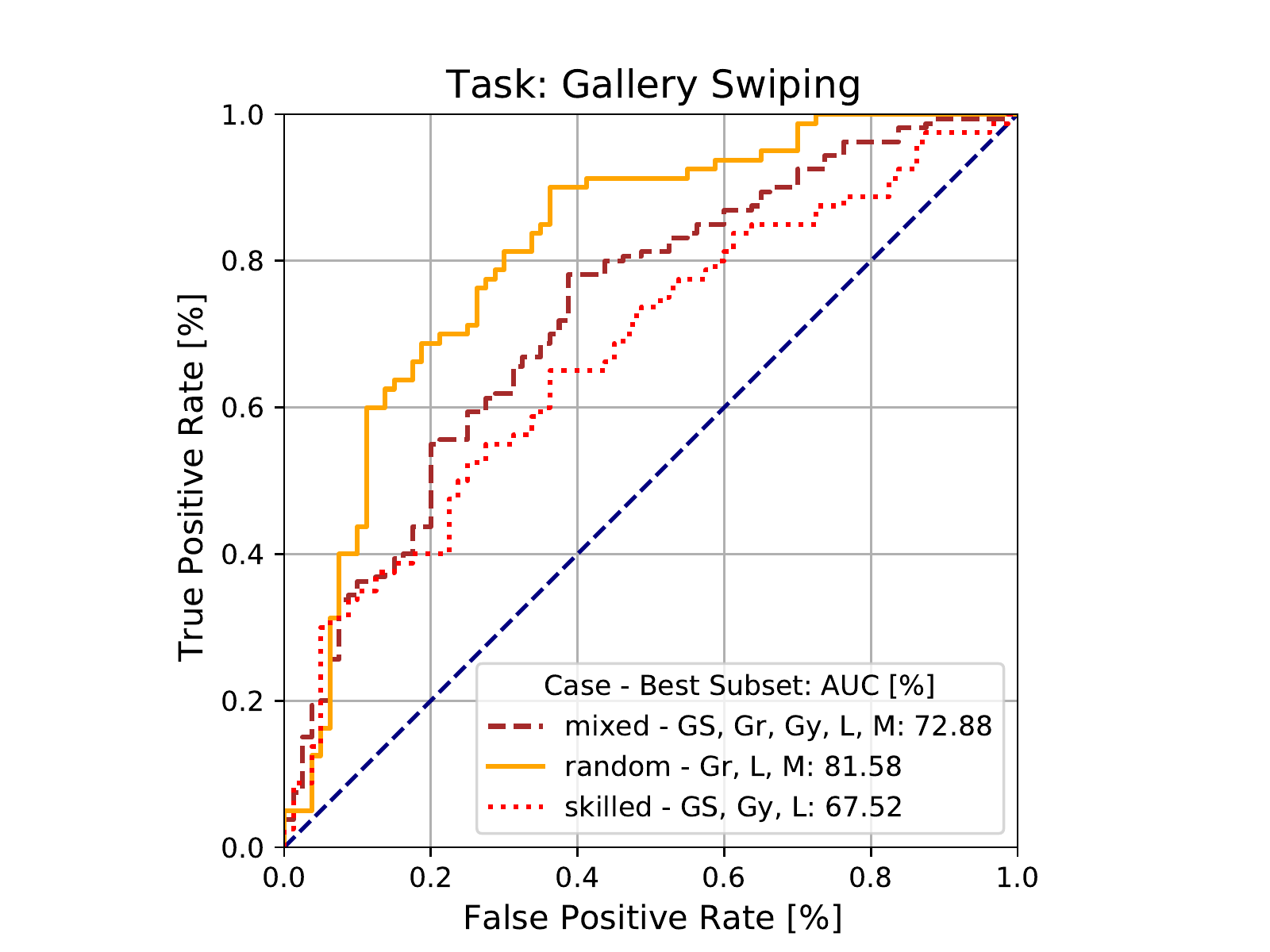}
\endminipage
\minipage{0.5\textwidth}
\adjincludegraphics[width=\linewidth,trim={0 0 {.11\width} {.11\width}}]{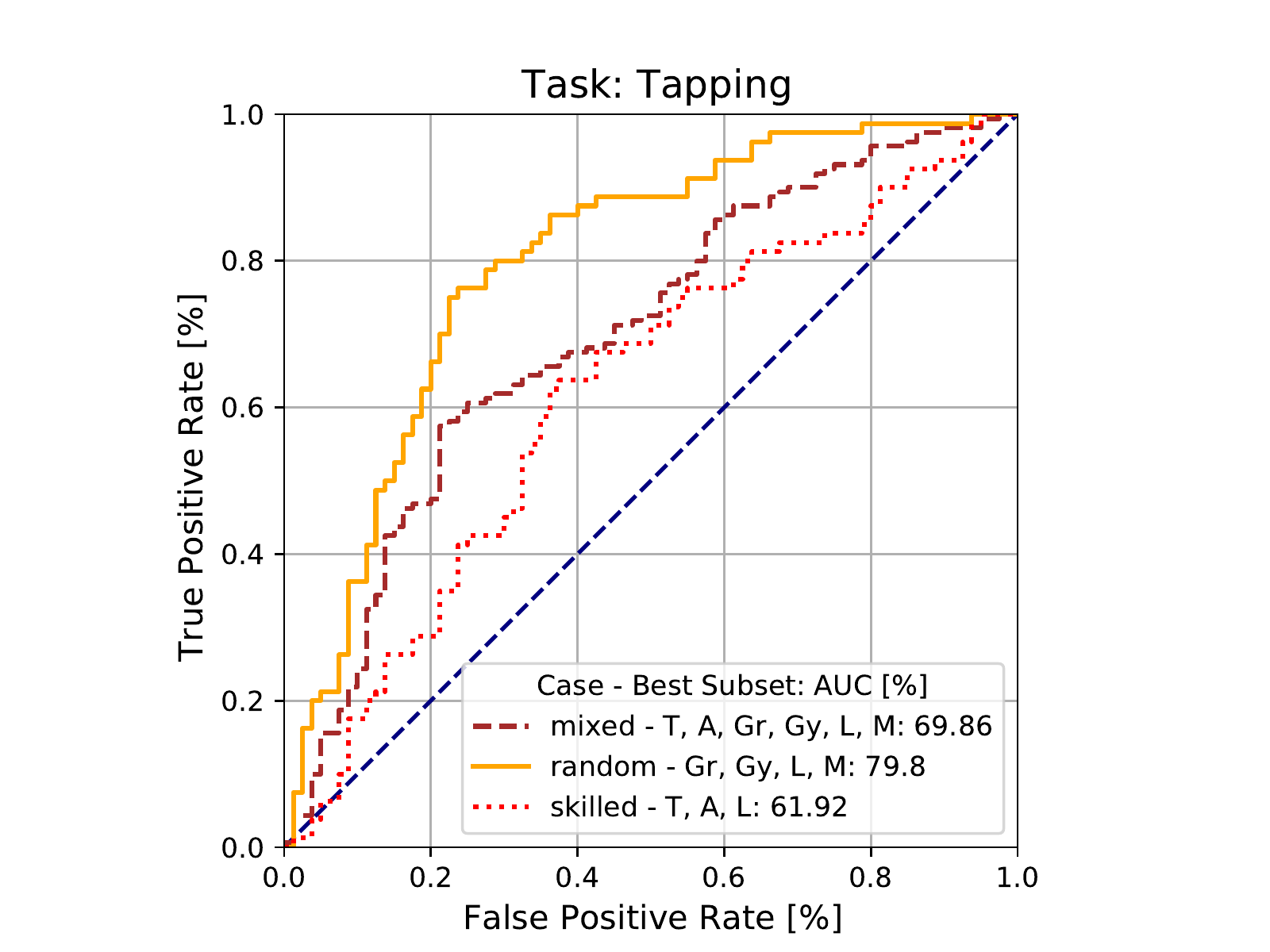}
\endminipage
% \vspace*{0.3cm}
    \captionsetup{width=\textwidth}
\caption{The ROC curves calculated for each of the tasks: keystroke, text reading, gallery swiping, tapping. Each ROC curves shows the best results for the random, skilled, and the mixed distributions of impostor data. The AUC value is also indicated in the legend.}
\vspace*{-0.5cm}
\end{figure*}

\section{Conclusions}
\label{sec:Conclusions}
This article has focused on an analysis of individual and multimodal behavioral biometric traits suitable for the application of mobile continuous authentication. 
The described work is included in a line of research that aims at providing a common ground for future research in the field of mobile behavioral biometrics, by addressing the lack of uniformity across different recently proposed studies, thus contributing to the advancement of the state of the art. To fill in this gap, we have taken several actions: \textit{(i)} we have presented BehavePassDB\footnote{\texttt{\url{https://github.com/BiDAlab/MobileB2C_BehavePassDB/}}}, a new publicly available database of wide ranging mobile interaction data, collected in an unsupervised scenario. The novel aspect of this database is the possibility to evaluate the effectiveness of developed systems considering the case of several users using their own device (random impostor scenario), and having impostors using the same device as the genuine users (skilled impostor scenario). \textit{(ii)} We have designed a standard experimental protocol, i. e., divided the database into development and evaluation sets, created the rules for the comparison of the genuine and impostor distributions, implemented the anonymization of the evaluation set, and adopted a popular metric in the field (AUC). \textit{(iii)} We have proposed MobileB2C\footnote{\texttt{\url{https://sites.google.com}}}, a competition at the International Joint Conference of Biometrics9 (IJCB) 2022 in order to evaluate systems developed by different research groups worldwide on the same dataset. \textit{(iv)} We have carried out a first benchmark of BehavePassDB\footnote{\texttt{\url{http://www.ijcb2022.org/}}}, considering several modalities such as touchscreen and background sensor data. For every individual modality, a separate LSTM RNN with triplet loss, and fusion at score level, has been considered. The current article will serve as the baseline for a ongoing competition which is being announced. Our experimental results show that the best performing source is the keystroke dynamics for touchscreen data, and the magnetometer and the gravity sensor for background sensors. Nevertheless, the discriminative ability of the system is significantly enhanced by the fusion, typically reaching a 80\%-87\% AUC range in the random impostor case and 62\%-69\% AUC in the skilled impostor case. It appears clear that the learned features in the random impostor case are not robust in the skilled impostor scenario.
\par Finally, we opted for a data collection approach in which the user interaction data are acquired during dense gesture-based dedicated sessions to maximize the amount of biometric information acquired. Consequently, it would be interesting, for future work, to assess possible system performance improvements with a larger database including more subjects both in training and in testing. Additionally, the generation of synthetic data, which proved to be a powerful tool in related fields, could be investigated \cite{tolosana2021deepwritesyn}.
”

\section*{Acknowledgment}

This project has received funding from the European Union’s Horizon 2020 research and innovation programme under the Marie Skłodowska-Curie grant agreement No 860315, and from Orange Labs. R. Vera-Rodriguez, R. Tolosana, and A. Morales are also supported by INTER-ACTION (PID2021-126521OB-I00 MICINN/FEDER). 

\bibliographystyle{IEEEtran}
\bibliography{bib}

\begin{thebibliography}{10}
\providecommand{\url}[1]{#1}
\csname url@samestyle\endcsname
\providecommand{\newblock}{\relax}
\providecommand{\bibinfo}[2]{#2}
\providecommand{\BIBentrySTDinterwordspacing}{\spaceskip=0pt\relax}
\providecommand{\BIBentryALTinterwordstretchfactor}{4}
\providecommand{\BIBentryALTinterwordspacing}{\spaceskip=\fontdimen2\font plus
\BIBentryALTinterwordstretchfactor\fontdimen3\font minus
  \fontdimen4\font\relax}
\providecommand{\BIBforeignlanguage}[2]{{%
\expandafter\ifx\csname l@#1\endcsname\relax
\typeout{** WARNING: IEEEtran.bst: No hyphenation pattern has been}%
\typeout{** loaded for the language `#1'. Using the pattern for}%
\typeout{** the default language instead.}%
\else
\language=\csname l@#1\endcsname
\fi
#2}}
\providecommand{\BIBdecl}{\relax}
\BIBdecl

\bibitem{WANG2020107118}
C.~Wang, Y.~Wang, Y.~Chen, H.~Liu, and J.~Liu, ``{User authentication on mobile
  devices: Approaches, threats and trends},'' \emph{Computer Networks}, vol.
  170, p. 107118, 2020.

\bibitem{marcel2019handbook}
S.~Marcel, M.~S. Nixon, J.~Fierrez, and N.~Evans, \emph{Handbook of Biometric
  Anti-Spoofing: Presentation Attack Detection}.\hskip 1em plus 0.5em minus
  0.4em\relax {2nd Ed., Springer}, 2019.

\bibitem{rathgeb2022handbook}
C.~Rathgeb, R.~Tolosana, R.~Vera-Rodriguez, and C.~Busch, ``Handbook of digital
  face manipulation and detection: From deepfakes to morphing attacks,'' 2022.

\bibitem{8998358}
R.~Tolosana, R.~Vera-Rodriguez, J.~Fierrez, and J.~Ortega-Garcia,
  ``{BioTouchPass2: Touchscreen Password Biometrics Using Time-Aligned
  Recurrent Neural Networks},'' \emph{IEEE Trans. on Information Forensics and
  Security}, vol.~15, pp. 2616--2628, 2020.

\bibitem{7503170}
V.~M. Patel, R.~Chellappa, D.~Chandra, and B.~Barbello, ``Continuous user
  authentication on mobile devices: Recent progress and remaining challenges,''
  \emph{IEEE Signal Processing Magazine}, vol.~33, no.~4, pp. 49--61, 2016.

\bibitem{stragapede2022prl}
G.~Stragapede, R.~Vera-Rodriguez, R.~Tolosana, A.~Morales, A.~Acien, and
  G.~Le~Lan, ``Mobile behavioral biometrics for passive authentication,''
  \emph{Pattern Recognition Letters}, 2022.

\bibitem{delgadosantos2021survey}
P.~Delgado-Santos, G.~Stragapede, R.~Tolosana, R.~Guest, F.~Deravi, and
  R.~Vera-Rodriguez, ``A survey of privacy vulnerabilities of mobile device
  sensors,'' \emph{ACM Computing Surveys}, 2022.

\bibitem{rasmussen15NDSS}
S.~Eberz, K.~B. Rasmussen, V.~Lenders, and I.~Martinovic, ``{Preventing
  Lunchtime Attacks: Fighting Insider Threats With Eye Movement Biometrics},''
  in \emph{The Network and Distributed System Security Symposium}, 2015.

\bibitem{Acien2020b}
A.~Acien, A.~Morales, J.~Fierrez, R.~Vera-Rodriguez, and O.~Delgado-Mohatar,
  ``{BeCAPTCHA: Behavioral bot detection using touchscreen and mobile sensors
  benchmarked on HuMIdb},'' \emph{Engineering Applications of Artificial
  Intelligence}, vol.~98, p. 104058, 2021.

\bibitem{Zhu2019}
T.~Zhu, Z.~Qu, H.~Xu, J.~Zhang, Z.~Shao, Y.~Chen, S.~Prabhakar, and J.~Yang,
  ``{RiskCog: Unobtrusive real-time user authentication on mobile devices in
  the wild},'' \emph{IEEE Transactions on Mobile Computing}, vol.~19, no.~2,
  pp. 466--483, 2019.

\bibitem{das2015exploring}
A.~Das and \textit{et al.}, ``Exploring ways to mitigate sensor-based
  smartphone fingerprinting,'' \emph{arXiv{:\texttt{1503.01874}}}, 2015.

\bibitem{Neverova2016}
N.~{Neverova}, C.~{Wolf}, G.~{Lacey}, L.~{Fridman}, D.~{Chandra},
  B.~{Barbello}, and G.~{Taylor}, ``{Learning Human Identity From Motion
  Patterns},'' \emph{IEEE Access}, vol.~4, pp. 1810--1820, 2016.

\bibitem{Deb2019}
D.~Deb, A.~Ross, A.~K. Jain, K.~Prakah-Asante, and K.~V. Prasad, ``{Actions
  Speak Louder Than (Pass)Words: Passive Authentication of Smartphone* Users
  via Deep Temporal Features},'' in \emph{Proc. 2019 Intl. Conf. on
  Biometrics}, 2019.

\bibitem{Abuhamad}
M.~Abuhamad, T.~Abuhmed, D.~Mohaisen, and D.~Nyang, ``{AUToSen:
  Deep-Learning-Based Implicit Continuous Authentication Using Smartphone
  Sensors},'' \emph{IEEE Internet of Things Journal}, vol.~7, no.~6, pp.
  5008--5020, 2020.

\bibitem{2020_CDS_HCIsmart_Acien}
A.~Acien, A.~Morales, R.~Vera-Rodriguez, and J.~Fierrez, ``{Smartphone Sensors
  For Modeling Human-Computer Interaction: General Outlook And Research
  Datasets For User Authentication},'' in \emph{Proc. IEEE Intl. Workshop on
  Consumer Devices and Systems}, 2020.

\bibitem{Palin_AaltoDBMobile19}
K.~Palin, A.~M. Feit, S.~Kim, P.~O. Kristensson, and A.~Oulasvirta, ``{How Do
  People Type on Mobile Devices? Observations from a Study with 37,000
  Volunteers},'' in \emph{Proc. of the 21st Intl. Conf. on Human-Computer
  Interaction with Mobile Devices and Services}, 2019.

\bibitem{9539873}
A.~Acien, A.~Morales, J.~V. Monaco, R.~Vera-Rodriguez, and J.~Fierrez,
  ``Typenet: Deep learning keystroke biometrics,'' \emph{IEEE Transactions on
  Biometrics, Behavior, and Identity Science}, vol.~4, no.~1, pp. 57--70, 2022.

\bibitem{stragapede2022iwbf}
{G. Stragapede, R. Vera-Rodriguez, R. Tolosana, A. Morales, A. Acien, and G. Le
  Lan}, ``Mobile passive authentication through touchscreen and background
  sensor data,'' in \emph{Proc. of the IEEE Intl. Workshop on Forensics and
  Biometrics}, 2022.

\bibitem{Mahbub2016}
U.~Mahbub, S.~Sarkar, V.~M. Patel, and R.~Chellappa, ``{Active user
  authentication for smartphones: A challenge data set and benchmark
  results},'' in \emph{Proc. of the 2016 IEEE 8th Intl. Conf. on Biometrics
  Theory, Applications and Systems}, 2016.

\bibitem{Shepard2011}
C.~Shepard, A.~Rahmati, C.~Tossell, L.~Zhong, and P.~Kortum, ``{LiveLab}:
  measuring wireless networks and smartphone users in the field,'' \emph{ACM
  SIGMETRICS Performance Evaluation Review}, vol.~38, no.~3, pp. 15--20, 2011.

\bibitem{Frank2012}
M.~Frank, R.~Biedert, E.~Ma, I.~Martinovic, and D.~Song, ``{Touchalytics: On
  the applicability of touchscreen input as a behavioral biometric for
  continuous authentication},'' \emph{IEEE transactions on information
  forensics and security}, vol.~8, no.~1, pp. 136--148, 2012.

\bibitem{Serwadda2013}
A.~Serwadda, V.~V. Phoha, and Z.~Wang, ``Which verifiers work?: A benchmark
  evaluation of touch-based authentication algorithms,'' in \emph{Proc. of the
  2013 IEEE 6th Intl. Conf. on Biometrics: Theory, Applications and Systems},
  2013.

\bibitem{Zhang2015}
H.~Zhang, V.~M. Patel, M.~Fathy, and R.~Chellappa, ``Touch gesture-based active
  user authentication using dictionaries,'' in \emph{Proc. of the 2015 IEEE
  Winter Conf. on Applications of Computer Vision}, 2015.

\bibitem{Feng2014}
T.~Feng, J.~Yang, Z.~Yan, E.~M. Tapia, and W.~Shi, ``{Tips: Context-aware
  implicit user identification using touch screen in uncontrolled
  environments},'' in \emph{Proc. of the 15th Workshop on Mobile Computing
  Systems and Applications}, 2014.

\bibitem{Saevanee2015}
H.~Saevanee, N.~Clarke, S.~Furnell, and V.~Biscione, ``Continuous user
  authentication using multi-modal biometrics,'' \emph{Computers \& Security},
  vol.~53, pp. 234--246, 2015.

\bibitem{Neal2015}
T.~J. Neal, D.~L. Woodard, and A.~D. Striegel, ``{Mobile device application,
  bluetooth, and Wi-Fi usage data as behavioral biometric traits},'' in
  \emph{Proc. of the 2015 IEEE 7th Intl. Conf. on Biometrics Theory,
  Applications and Systems}, 2015.

\bibitem{Wu2015}
J.~Wu and Z.~Chen, ``An implicit identity authentication system considering
  changes of gesture based on keystroke behaviors,'' \emph{International
  Journal of Distributed Sensor Networks}, vol.~11, no.~6, p. 470274, 2015.

\bibitem{Murmuria2015}
R.~Murmuria, A.~Stavrou, D.~Barbar{\'a}, and D.~Fleck, ``Continuous
  authentication on mobile devices using power consumption, touch gestures and
  physical movement of users,'' in \emph{Proc. of the Intl. Symp. on Recent
  Advances in Intrusion Detection}, 2015.

\bibitem{Sitova2015}
Z.~Sitov{\'a}, J.~{\v{S}}ed{\v{e}}nka, Q.~Yang, G.~Peng, G.~Zhou, P.~Gasti, and
  K.~S. Balagani, ``{HMOG: New behavioral biometric features for continuous
  authentication of smartphone users},'' \emph{IEEE Transactions on Information
  Forensics and Security}, vol.~11, no.~5, pp. 877--892, 2015.

\bibitem{Lu2015}
L.~Lu and Y.~Liu, ``Safeguard: User reauthentication on smartphones via
  behavioral biometrics,'' \emph{IEEE Transactions on Computational Social
  Systems}, vol.~2, no.~3, pp. 53--64, 2015.

\bibitem{Coakley2016}
M.~J. {Coakley}, J.~V. {Monaco}, and C.~C. {Tappert}, ``Keystroke biometric
  studies with short numeric input on smartphones,'' in \emph{Proc. of the 2016
  IEEE 8th Intl. Conf. on Biometrics Theory, Applications and Systems}, 2016.

\bibitem{Kumar2016}
R.~Kumar, V.~V. Phoha, and A.~Serwadda, ``Continuous authentication of
  smartphone users by fusing typing, swiping, and phone movement patterns,'' in
  \emph{Proc. of the 2016 IEEE 8th Intl. Conf. on Biometrics Theory,
  Applications and Systems}, 2016.

\bibitem{Lee2017}
W.-H. Lee and R.~B. Lee, ``Implicit smartphone user authentication with sensors
  and contextual machine learning,'' in \emph{Proc. of the 2017 47th Annual
  IEEE/IFIP Intl. Conf. on Dependable Systems and Networks}, 2017.

\bibitem{Zhu2017}
H.~Zhu, J.~Hu, S.~Chang, and L.~Lu, ``{ShakeIn: Secure user authentication of
  smartphones with single-handed shakes},'' \emph{IEEE transactions on mobile
  computing}, vol.~16, no.~10, pp. 2901--2912, 2017.

\bibitem{AlKork2017}
S.~K. Al~Kork, I.~Gowthami, X.~Savatier, T.~Beyrouthy, J.~A. Korbane, and
  S.~Roshdi, ``Biometric database for human gait recognition using wearable
  sensors and a smartphone,'' in \emph{Proceedings of the 2017 2nd Intl. Conf.
  on Bio-engineering for Smart Technologies}, 2017.

\bibitem{Li2018}
G.~Li and P.~Bours, ``{Studying WiFi and accelerometer data based
  authentication method on mobile phones},'' in \emph{Proc. of the 2018 2nd
  Intl. Conf. on Biometric Engineering and Applications}, 2018.

\bibitem{Amini2018}
S.~Amini, V.~Noroozi, A.~Pande, S.~Gupte, P.~S. Yu, and C.~Kanich, ``Deepauth:
  A framework for continuous user re-authentication in mobile apps,'' in
  \emph{Proc. of the 27th ACM Intl. Conf. on Information and Knowledge
  Management}, 2018.

\bibitem{TOLOSANA2022108609}
{R. Tolosana, R. Vera-Rodriguez \textit{et al.}}, ``{SVC-onGoing: Signature
  verification competition},'' \emph{Pattern Recognition}, vol. 127, p. 108609,
  2022.

\bibitem{Deepsigntolosana}
{R. Tolosana, R. Vera-Rodriguez, J. Fierrez and J. Ortega-Garcia}, ``Deepsign:
  Deep on-line signature verification,'' \emph{IEEE Transactions on Biometrics,
  Behavior, and Identity Science}, vol.~3, no.~2, pp. 229--239, 2021.

\bibitem{tolosana2021deepwritesyn}
R.~Tolosana, P.~Delgado-Santos, A.~Perez-Uribe, R.~Vera-Rodriguez, J.~Fierrez,
  and A.~Morales, ``Deepwritesyn: On-line handwriting synthesis via deep
  short-term representations,'' in \emph{Proc. AAAI Conference on Artificial
  Intelligence}, 2021.

\end{thebibliography}

\end{document}